\documentclass[10pt,letterpaper]{article}

\usepackage{ccn}
\usepackage{pslatex}
\usepackage{apacite}
\usepackage[round]{natbib}
\usepackage{booktabs}
\usepackage{graphicx}
\usepackage{svg}
\usepackage{hyperref}
\usepackage{cleveref}
\usepackage{microtype}

\usepackage{footnote}
\usepackage[flushmargin]{footmisc} 
\renewcommand{\thefootnote}{\arabic{footnote}}
\newcommand{\astfootnote}[1]{%
\let\oldthefootnote=\thefootnote%
\setcounter{footnote}{0}%
\renewcommand{\thefootnote}{\fnsymbol{footnote}}%
\footnote{#1}%
\let\thefootnote=\oldthefootnote%
}
\newcommand\blfootnote[1]{%
  \begingroup
  \renewcommand\thefootnote{}\footnote{#1}%
  \addtocounter{footnote}{-1}%
  \endgroup
}

\usepackage{todonotes}
\usepackage{comment}

\title{Does the brain represent words?\\An evaluation of brain decoding studies of language understanding}

\author{{\large {\bf Jon Gauthier} (jon@gauthiers.net) }
  and {\large {\bf Anna Ivanova}\thanks{Authors contributed equally.} (annaiv@mit.edu) } \\
  Department of Brain and Cognitive Sciences,
  Massachusetts Institute of Technology \\
  }

\date{}

\begin{document}

\maketitle

\section{Abstract}
{
\bf
Language decoding studies have identified word representations which can be used to predict brain activity in response to novel words and sentences \citep{anderson2016predicting, pereira2018toward}. The unspoken assumption of these studies is that, during processing, linguistic information is transformed into some shared semantic space, and those semantic representations are then used for a variety of linguistic and non-linguistic tasks. We claim that current studies vastly underdetermine the content of these representations, the algorithms which the brain deploys to produce and consume them, and the computational tasks which they are designed to solve. We illustrate this indeterminacy with an extension of the sentence-decoding experiment of \citet{pereira2018toward}, showing how standard evaluations fail to distinguish between language processing models which deploy different mechanisms and which are optimized to solve very different tasks. We conclude by suggesting changes to the brain decoding paradigm which can support stronger claims of neural representation.
}
\begin{quote}
\small
\textbf{Keywords:}
brain decoding; language; representation
\end{quote}

\blfootnote{Code available at \href{https://github.com/hans/nn-decoding}{\texttt{github.com/hans/nn-decoding}}.}

Ever since the seminal word decoding paper by \citet{mitchell2008predicting} researchers have attempted to come up with a feature space that best captures semantic aspects of language processing in the brain. Many of those studies refer to a neural activity pattern evoked by a particular stimulus as a ''semantic representation'' of that stimulus. Their goal is to isolate a set of stimulus features which best capture brain activity. This paper will argue that the problem the field is trying to solve is ill-defined: such talk of representation is meaningless unless one also specifies the brain mechanisms utilizing those representations and the task they are designed to solve.

The original decoding study by \citet{mitchell2008predicting} demonstrated that a model derived from the co-occurrence patterns of 25 sensorimotor verbs performed above chance both when predicting fMRI responses to novel nouns and when selecting a noun matching a previously unseen fMRI image. Since then, others have proposed alternate word feature models, based either on behavioral ratings \citep[e.g.,][]{binder2016toward} or on distributional statistics \citep[e.g.,][]{devereux2010using, murphy2012selecting}. A direct comparison of some of those models revealed that their relative performance differs for encoding and decoding tasks and even varies from subject to subject \citep{abnar2018}. Both the experience-based and the distributional approaches have recently been extended to sentence-level decoding \citep{anderson2016predicting, pereira2018toward}, and have begun to use more sophisticated models accounting for context and grammatical information \citep{jain2018incorporating,anderson2018multiple}.

\subsection{What is a ``semantic representation?''} The main goal of a language decoding study is to derive a set of stimulus-specific linguistic features and estimate how it is associated with brain activity. If we find that brain activity patterns can predict these input features, we conclude that the brain activity contains a ``representation'' of the features. If those features reflect semantic properties of the stimulus, the brain activity pattern earns the label ``semantic representation.''
Such representational claims are dangerously weak: they wildly overgenerate, leading us to award the label of ``representation'' to brain activity evoked by any arbitrary aspect of the stimulus, so long as it has some vague relation to the stimulus ``meaning''.

In what sense does the above claim overgenerate? Consider the decoding studies in the line of \citet{pereira2018toward}, which use fMRI data from subjects reading a sentence to predict embedding vectors of the words in that sentence. Embedding-based representations have been shown to capture many different aspects of words, from simple features like word frequency and logical relationships such as hypernymy \citep{fu2014learning} to arbitrarily complex features of both syntax and semantics \citep{mikolov2013linguistic}. While \citeauthor{pereira2018toward} claim that their decoder in this task has learned to read out ``linguistic meaning'' from the brain, we could just as well claim that the decoder has captured ``elements of syntax'' or ``hypernymy relations.''

People are obviously doing more than reasoning about syntax or hypernymy relations when reading a sentence. But such decoding results are consistent with \emph{all} of these claims. In short, the decoding approach underdetermines the actual nature and function of neural computations.

Representations do not exist in a vacuum. They are \emph{computed} by specific neural mechanisms, and are \emph{consumed} by other systems for the purpose of producing behavior. Claims of ``semantic representation'' without an accompanying description of such mechanisms are dangerously underspecified. In light of these conceptual issues, philosophers concerned with neural representation have concluded that representational claims are incomplete without a description of the associated producing and consuming mechanisms \citep{papineau1992reality,dretske1995naturalizing}.


We demonstrate this incompleteness by drawing on neural network models from natural language processing which produce intermediate representations of input sentences. These representations are importantly \emph{not} produced in a vacuum: they are optimized to best help their corresponding model succeed in some downstream task. We select a wide range of models, ranging from machine translation to image-caption retrieval. We re-run the experiment of \citet{pereira2018toward} by learning decoder models that map the subjects' fMRI data recorded while listening to a sentence to a neural net representation of that sentence. Representations from all but one of the neural nets perform above chance in our decoding trials, suggesting that brain activity captures some representational elements within \emph{each} of these task-specific models.

Of course, we do not take this result to imply that subjects are jointly solving machine translation and image-caption retrieval tasks while reading sentences. We instead take it to imply that the current decoding evaluation method is too weak: it underdetermines both the task being solved by the subject and the neural mechanisms deployed to solve that task. We conclude by proposing a move away from target models based on simple feature-response correlations toward target models which make explicit commitments to the operant task and neural mechanisms designed to solve that task.

\section{Experiment}

\begin{figure}[bth]
    \centering
    \includegraphics[width=.85\linewidth]{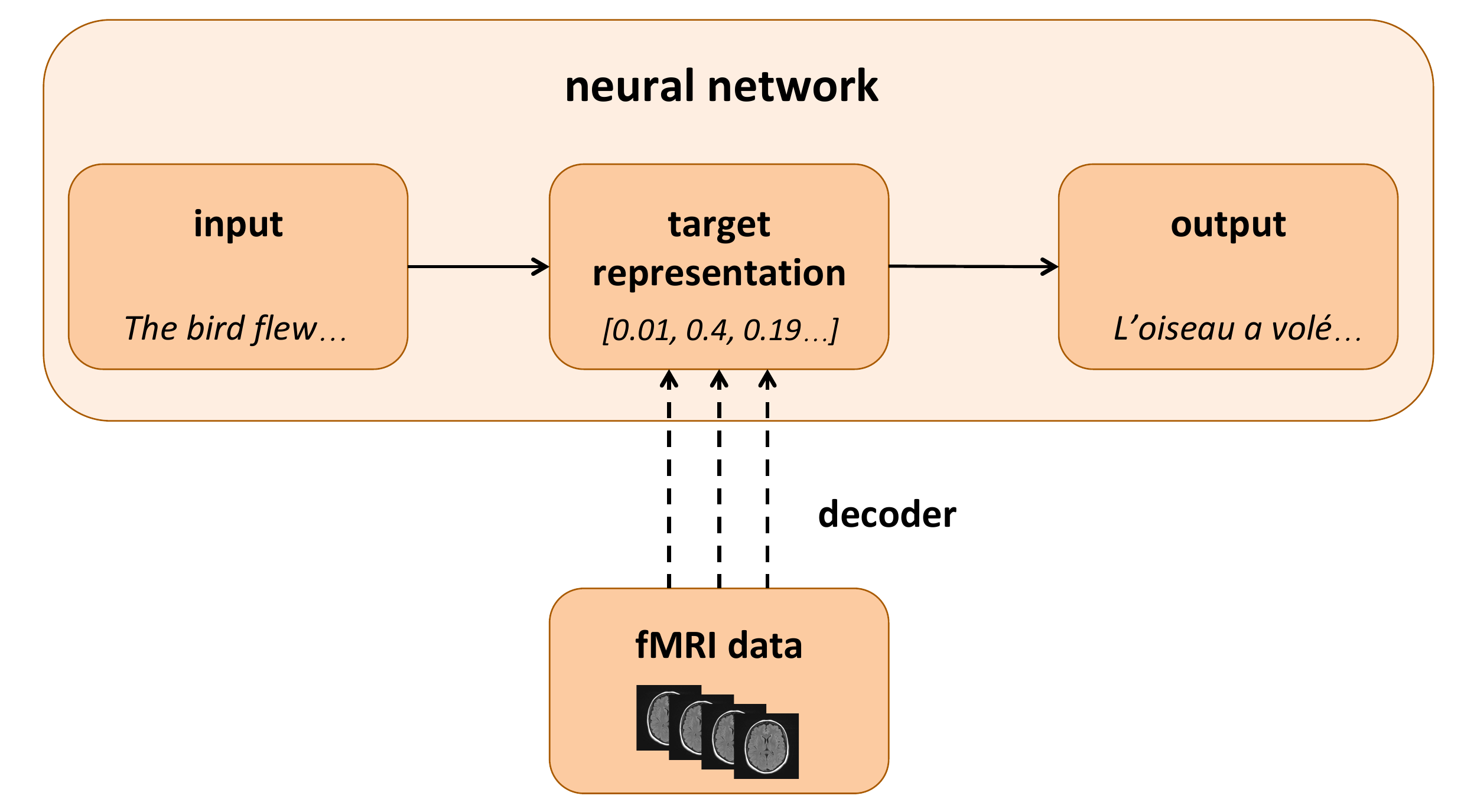}
    \caption{We train decoder models to predict intermediate representations of neural network models for natural language processing tasks. A model performing machine translation is shown here.}
    \label{fig:framework}
\end{figure}

We perform a decoding experiment, mapping from the brain activations of subjects to the intermediate representations of various neural network models for natural language processing. We use brain imaging data from \citet[experiment 2]{pereira2018toward}, who presented subjects with 384 sentences, each involving one of 180 different common words. 
Each subject was instructed to carefully read each sentence, presented one at a time. The subjects' brain activity in response to each sentence was saved to be used as inputs for decoding trials.\footnote{For more information on imaging methods and data preprocessing, please see \citet{pereira2018toward}.}

\paragraph{Sentence representations}

\begin{table}[thb]
    \centering
    \resizebox{\linewidth}{!}{
    \begin{tabular}{clll}
        \toprule
        & Name & Task & Architecture \\
        \midrule
        1 & GloVe & Distributional modeling & GloVe \citep{pennington2014glove} \\
        2 & skipthought & Language modeling & LSTM \citep{kiros2015skip} \\
        3 & InferSent & Natural language inference & LSTM \citep{conneau2017supervised} \\
        4 & DisSent & Discourse understanding & BiLSTM \citep{nie2017dissent} \\
        5 & ConvS2S & Machine translation & CNN+attention \citep{gehring2017convs2s} \\
        6 & order & Image caption retrieval & LSTM \citep{vendrov2015order} \\
        7 & IMDB & Sentiment & LSTM (vanilla) \\
        \bottomrule
    \end{tabular}
    }
    \caption{Models used to compute target sentence representations in our experiments.}
    \label{tbl:models}
\end{table}

\Cref{tbl:models} summarizes the neural network models used to produce target representations in our experiments. We select a large class of models which share a single architectural feature: each model takes a sentence as input, computes a high-dimensional vector representing that sentence, and uses this representation to make task-specific decisions. Each model is trained to produce representations which help it optimally perform these task-specific decisions. It is important to note that these models are optimized for vastly different tasks: the machine translation model, for example, creates a representation of an English input sentence in order to output its French translation; the image caption retrieval model computes a representation of a sentence in order to match the sentence to the most relevant images in a test set.

\paragraph{Decoding}

Let $r(x_i, s_j)$ be the fMRI activity recorded after a subject $s_j$ observes a sentence $x_i$, and $r(x_i, m_k)$ be the sentence representation output by model $m_k$ given sentence $x_i$ as input. For each subject $s_j$ and model $m_k$, we learn a ridge regression which predicts the model's sentence representation $r(x_i, m_k)$ given the subject's fMRI response $r(x_i, s_j)$. The training loss for a decoder $\theta_{jk}$ mapping from brain images of subject $s_j$ to representations of model $m_k$ is as follows:
\begin{equation}
J(s_j, m_k) = \frac{1}{|x|} \sum_{x_i} \left( || r(x_i, s_j)^T \theta_{jk} - r(x_i, m_k) ||^2_2 \right) + \alpha || \theta_{jk} ||^2_2
\end{equation}

\noindent{}where $\alpha$ is a regularization hyperparameter. For each subject--model pairing, we train decoder weights $\theta_{jk}$ on the 384 sentences used by \citeauthor{pereira2018toward} and the associated brain images.

\paragraph{Evaluation method}

We evaluate the learned decoders using the \emph{mean average rank} metric from \citeauthor{pereira2018toward} Given an input brain image $r(x_i, s_j)$ of a subject $s_j$ observing sentence $x_i$, we use the decoder parameters $\theta_{jk}$ to predict a model representation $\hat r(x_i, m_k)$. We next calculate the cosine distance between this predicted representation and the known representations of the 384 sentences. The rank score of a prediction is the position of the encoding of the actual sentence $x_i$ in the ranked list.\footnote{A perfect model would yield a rank score of 0 on every input; a model making random predictions would yield a rank score of $\frac 1 2 |x| = 192$ in our experiments.} To assign each decoder a single score, we calculate the average rank score of the decoder's predictions for all possible sentences. We take the mean of these average rank scores across subjects, yielding a single mean average rank score associated with the model representations:
\begin{equation}
MAR(m_k) = \frac{1}{|s|} \sum_{s_j} \frac{1}{|x|} \sum_{x_i} \textit{rank}(\hat r(x_i, m_k), r(x_i, m_k))
\end{equation}

\begin{figure}[thb]
    \centering
    \includegraphics[width=\linewidth]{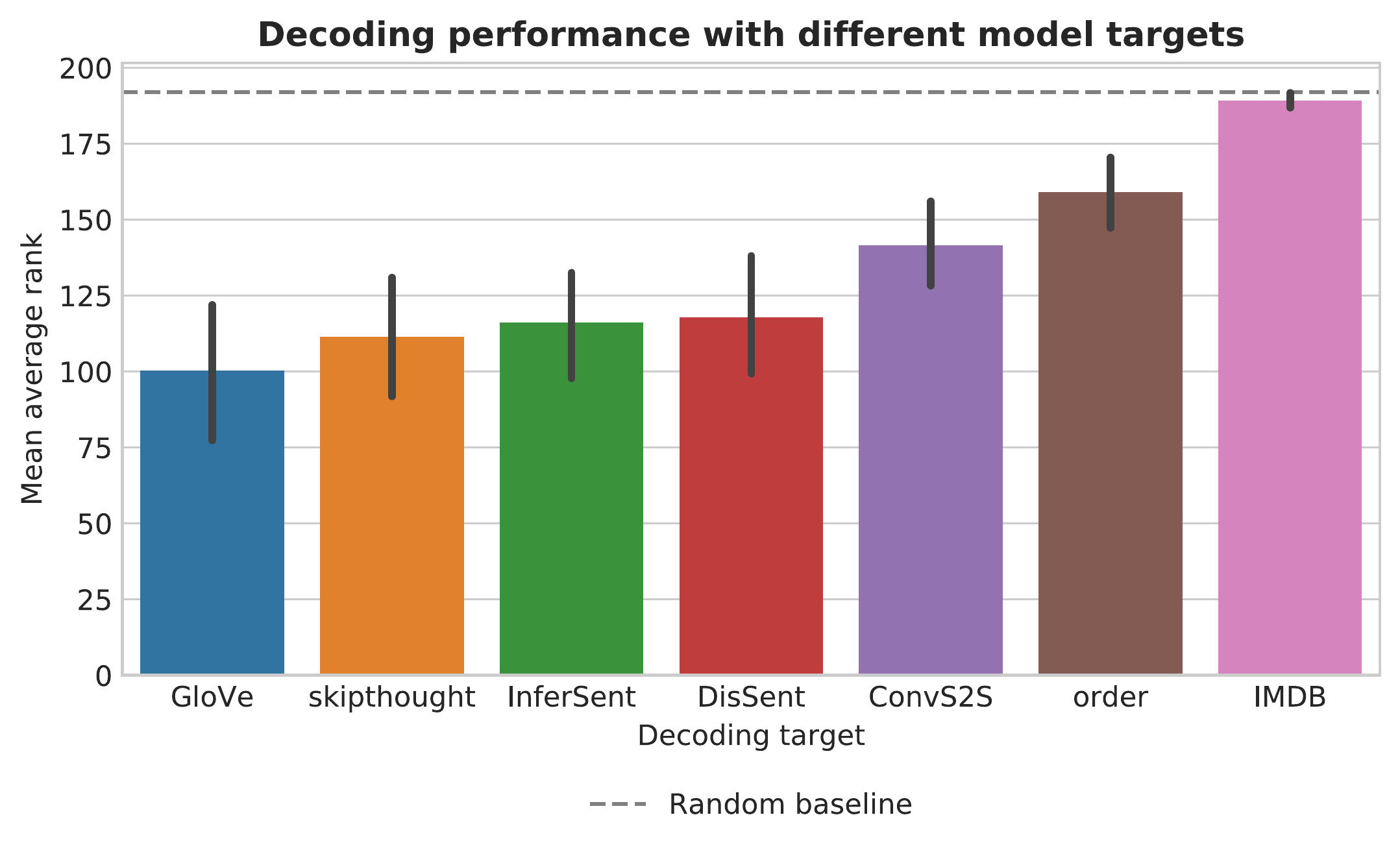}
    \caption{Reverse inference decoding underdetermines the task being solved. Mean average rank (MAR) metrics calculated on decoders trained to map between subjects' brain activity and target representations of models, each optimized to solve a different task. (Error bars denote bootstrap 95\% CI.)}
    \label{fig:mar}
\end{figure}

\begin{figure}[thb]
    \centering
    \includegraphics[width=0.95\linewidth]{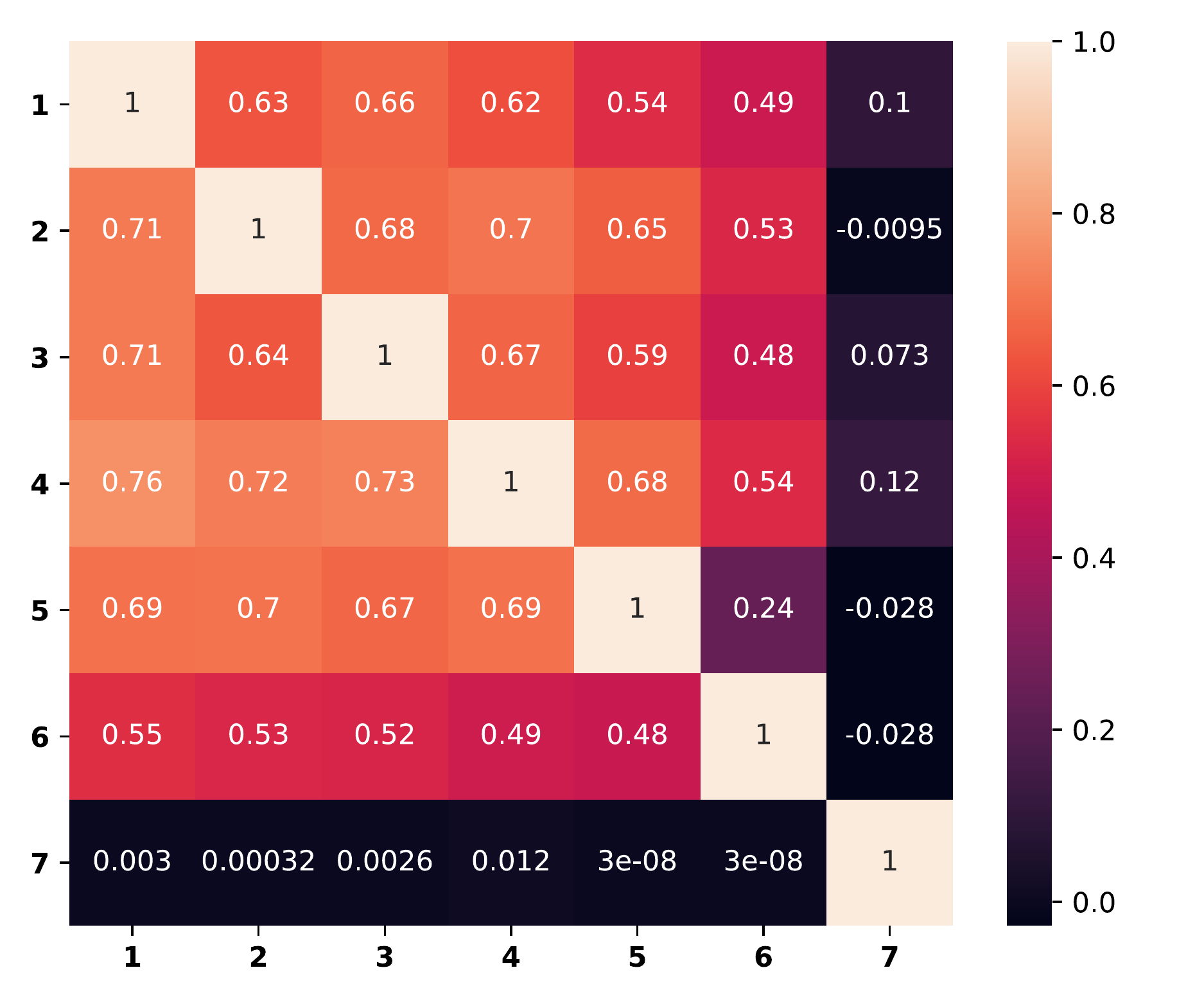}
    \caption{Heatmap describing different model representations used in our evaluation. Each cell describes the $r^2$ regression metric for mapping the representations of the row's model to representations of the column's model. (Numbers refer to rows of \Cref{tbl:models}, and are ordered as in \Cref{fig:mar}.)}
    \label{fig:heatmap}
\end{figure}

\paragraph{Results}

We compare the performance in our seven decoder trials, each of which attempts to map subject brain activity to the representations of one of seven target model representations. The results are shown in \Cref{fig:mar}. All decoders perform better than chance, except the decoder predicting activations of the sentiment analysis model. The four highest-performing models (achieving at best a MAR metric of $\sim$100) do not differ significantly in performance.\footnote{Not significant under a bootstrap comparison of 95\% confidence intervals of mean MAR.} Note that the most successful models in these trials were originally optimized to solve a variety of different tasks, from predicting the next word in a sentence to reasoning about the logical entailment relationships between sentences.

Are the highest-performing models performing well in our evaluation because they replicate \emph{overlapping} aspects of the subjects' brain activity, or because they incorporate \emph{complementary} features of brain activity? To answer this question, we analyze the similarity between each pair of model representations. For each model $m_i$, we learn a regression model mapping its representations to the representations of every other model $m_j \ne m_i$ for each of the 384 sentences in our experiment. \Cref{fig:heatmap} reports the $r^2$ metric of each pairwise regression evaluation (where row $i$ and column $j$ specify the $r^2$ metric resulting from mapping representations of model $i$ to those of model $j$). We can see that the observed pattern mostly agrees with results from \Cref{fig:mar}: representations that can be predicted by the baseline model representations can also be decoded more easily. The four models which yield the highest decoding performance in \Cref{fig:mar} are also relatively highly correlated between one another (\Cref{fig:heatmap}), suggesting that these models' representations share some underlying state also present in the subjects' brain images. Nevertheless, the present framework makes it hard to isolate the aspects of sentence representations that drive the decoders' performance.


\section{Discussion}

We find that state-of-the-art evaluation techniques used in decoding studies fail to distinguish between sentence representations drawn from models optimized for very different tasks. Decoders trained on representations from 5 out of 6 neural network models that we tested performed above chance, with three matching the performance of the baseline decoder trained used in the original study of \citet{pereira2018toward}.

The four best performing networks (originally trained to predict distributional information, entailment relations, or discourse structure) were all designed with the explicit purpose of producing task-transferable sentence representations, while the three networks which fare worse were originally trained in more narrowly defined tasks (machine translation, image captioning, and sentiment analysis). Our results suggest that the NLP community's quest for a universal sentence representation may actually be on the right track: these model representations appear to capture aspects of human sentence processing relatively well.\footnote{\citet{wang2018glue} evaluate how these representations transfer to different applied language tasks.} However, all that the neuroimaging community gains from such analyses is the knowledge that neural activity reflects a mechanism which is solving \emph{something} in the intersection of these diverse tasks.



If we want to better understand the way linguistic processing is realized in brain activity, we should adopt target models that are explicit about both the \emph{task} which they are optimizing and the downstream \emph{mechanisms} which operate on their representations. Some features might be used by multiple mechanisms (most of the neural network representations in our analysis are partially predictive of each other, despite being suited for different tasks), while others might be exclusive to one or two. Below, we outline some approaches which we believe could support more interpretable language decoding studies.




\begin{description}
    \item[Commit to a specific mechanism and task.] In addition to claiming that brain activity can predict certain stimulus features, we should explicitly link those features with generating or consuming mechanisms. \citet{kay2008identifying} presented an encoding model that illustrates this approach. The model made an explicit mechanistic commitment, using Gabor wavelet functions to compute features of the image inputs. This design decision was based on prior knowledge that early visual cortices use Gabor-like filters to extract low-level image properties. Their success  does more than indicate that visual cortex ``represents images'': it provides evidence for a specific mechanistic claim about how those representations are produced.
    \item[Subdivide the input feature space.] Since language processing is complex, it is highly unlikely that brain activity evoked by a linguistic stimulus resides within some universal semantic space. Instead, the neural signal reflects the joint activity of multiple networks that process linguistic input using a wide variety of algorithms, each operating on its own representation of the input. One way to reflect this heterogeneity is by specifying several sets of input features, each of which captures a representation optimized for a particular task. \citet{wehbe2014simultaneously}, for example, labeled every word with a set of visual, syntactic, semantic, and discourse features. Evaluating the complete model, as well as models that included only one set of features, allowed them to determine whether brain activity patterns reflecting those feature sets overlap, as well as which regions work with which representations.
    \item[Explicitly measure explained variance.] \citet{naselaris2011encoding} show that decoding models cannot identify the full set of features that drive a response in a particular brain region. Encoding models, however, can be explicitly measured in their ability to explain the activity in a particular brain region \citep[see, e.g.,][]{wehbe2014simultaneously, anderson2018multiple}. Future work on language representation should evaluate the extent to which each model component can explain the overall fMRI response.

\end{description}

We have demonstrated that current decoding evaluation methods provide only indeterminate answers to the most important questions regarding neural representation: what mechanisms are responsible for producing and consuming such representations, and what task are the systems which consume these representations attempting to solve? Devising mechanistic models, breaking down the notion of semantic representation into sub-components, and explicitly testing mechanistic models against brain activity all appear to be a fruitful venues for future research in this domain.

\paragraph{Acknowledgements}
We thank Allen Nie and members of the MIT BCS Philosophy Circle for their comments and ideas.

\bibliographystyle{apacite}

\setlength{\bibleftmargin}{.125in}
\setlength{\bibindent}{-\bibleftmargin}
\def\bibfont{\footnotesize}

\bibliography{main}

\end{document}